\documentstyle[aaai]{article}
\begin{document}
\bibliographystyle{aaai}

\title{Some Advances in Transformation-Based Part of Speech
Tagging\thanks{This paper appears in the proceedings of the Twelfth
National Conference on Artificial
Intelligence (AAAI-94).  This research was supported by ARPA under
contract N00014-89-J-1332, monitored through the Office of Naval Research.}}
\author{Eric Brill \\
Spoken Language Systems Group \\ Laboratory for Computer
Science \\ Massachusetts Institute of Technology \\ Cambridge,
Massachusetts 02139 \\
brill@goldilocks.lcs.mit.edu}
\input{psfig}
\maketitle

\begin{abstract}

\begin{quote}
Most recent research in trainable part of speech taggers has explored
stochastic tagging.  While these taggers obtain high accuracy,
linguistic information is captured indirectly, typically in tens of
thousands of lexical and contextual probabilities.  In \cite{Brill92},
a trainable rule-based tagger was described that obtained performance
comparable to that of stochastic taggers, but captured relevant linguistic
information in a small number of simple non-stochastic rules.  In this
paper, we describe a number of extensions to this rule-based tagger.
First, we describe a method for expressing lexical relations in
tagging that stochastic taggers are currently unable to express.
Next, we show a rule-based approach to tagging unknown words.
Finally, we show how the tagger can be extended into a k-best tagger,
where multiple tags can be assigned to words in some cases of
uncertainty.
\end{quote}

\end{abstract}

\section{Introduction}

When automated part of speech tagging was initially explored
\cite{Klei63,Harr62}, people manually engineered rules for tagging,
sometimes with the aid of a corpus.  As large corpora became
available, it became clear that simple Markov-model based stochastic
taggers that were automatically trained could achieve high rates of
tagging accuracy \cite{Jeli85}.  Markov-model based taggers assign a
sentence the tag sequence that maximizes $Prob(\mbox{word}|\mbox{tag})
* Prob(\mbox{tag}|\mbox{previous n tags})$.  These probabilities can
be estimated directly from a manually tagged corpus.\footnote{One can
also estimate these probabilities without a manually tagged corpus,
using a hidden Markov model.  However, it appears to be the case that
directly estimating probabilities from even a very small manually
tagged corpus gives better results than training a hidden Markov model
on a large untagged corpus (see \cite{Merialdo91}).}  Stochastic
taggers have a number of advantages over the manually built taggers,
including obviating the need for laborious manual rule construction,
and possibly capturing useful information that may not have been
noticed by the human engineer. However, stochastic taggers have the
disadvantage that linguistic information is only captured indirectly,
in large tables of statistics. Almost all recent work in developing
automatically trained part of speech taggers has been on further
exploring Markov-model based tagging
\cite{Jeli85,Chur88,Dero88,DeMarcken90,Merialdo91,Cutt92,Kupiec92,Charniak93,We
ish93}.

In \cite{Brill92}, a trainable rule-based tagger is described that
achieves performance comparable to that of stochastic taggers.
Training this tagger is fully automated, but unlike trainable
stochastic taggers, linguistic information is encoded directly in a
set of simple non-stochastic rules.  In this paper, we describe some
extensions to this rule-based tagger.  These include a rule-based
approach to: lexicalizing the tagger, tagging unknown words, and
assigning the k-best tags to a word. All of these extensions, as well
as the original tagger, are based upon a learning paradigm called
transformation-based error-driven learning.  This learning paradigm
has shown promise in a number of other areas of natural language
processing, and we hope that the extensions to transformation-based
learning described in this paper can carry over to other domains of
application as well.\footnote{The programs described in this paper can
be obtained by contacting the author.}

\section{Transformation-Based Error-Driven Learning}

\begin{figure}
\centerline{\psfig{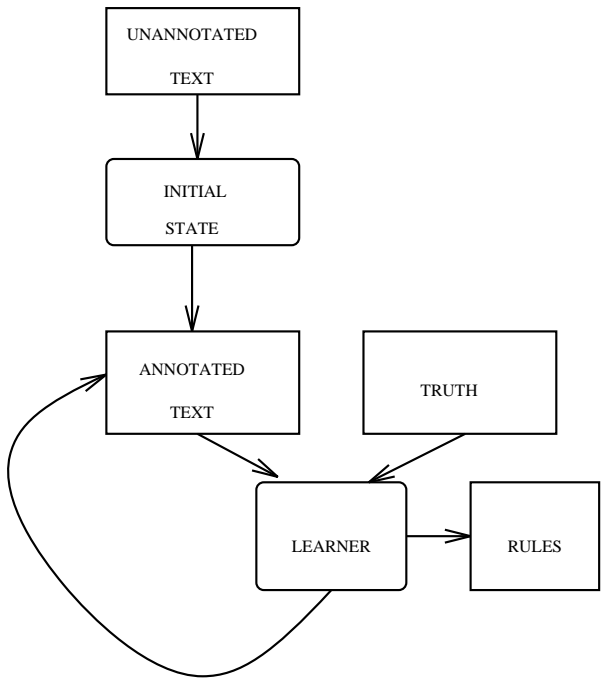}}
\caption{Transformation-Based Error-Driven Learning.}
\label{ded-learn}
\end{figure}

Transformation-based error-driven learning has been applied to a
number of natural language problems, including part of speech tagging,
prepositional phrase attachment disambiguation, and syntactic parsing
\cite{Brill92,Brill93,Brill93diss}.  A similar approach is being
explored for machine translation \cite{Su92}.  Figure \ref{ded-learn}
illustrates the learning process.  First, unannotated text is passed
through the initial-state annotator.  The initial-state annotator can
range in complexity from assigning random structure to assigning the
output of a sophisticated manually created annotator.  Once text has
been passed through the initial-state annotator, it is then compared
to the {\em truth},\footnote{As specified in a manually annotated
corpus.} and transformations are learned that can be applied to the
output of the initial state annotator to make it better resemble the
{\em truth}.

In all of the applications described in this paper, the following
greedy search is applied: at each iteration of learning, the
transformation is found whose application results in the {\em highest
score}; that transformation is then added to the ordered
transformation list and the training corpus is updated by applying the
learned transformation.  To define a specific application of
transformation-based learning, one must specify the following: (1) the
initial state annotator, (2) the space of transformations the learner is
allowed to examine, and (3) the scoring function for comparing the
corpus to the {\em truth} and choosing a transformation.

Once an ordered list of transformations is learned, new text can be
annotated by first applying the initial state annotator to it and then
applying each of the learned transformations, in order.

\section{An Earlier Tranformation-Based Tagger}

The original transformation-based tagger \cite{Brill92} works as
follows.  The initial state annotator assigns each word its most likely
tag as indicated in the training corpus.  The most likely tag for
unknown words is guessed based on a number of features, such as
whether the word is capitalized, and what the last three letters of
the word are.  The allowable transformation templates are:

\vspace{3mm}

Change tag {\bf a} to tag {\bf b} when:

\begin{enumerate}
\item The preceding (following) word is tagged {\em z}.
\item The word two before (after) is tagged {\em z}.
\item One of the two preceding (following) words is tagged {\em z}.
\item One of the three preceding (following) words is tagged {\em z}.
\item The preceding word is tagged {\em z} and the following  word is
tagged {\em w}.
\item The preceding (following) word is tagged  {\em z} and the word two
before (after) is tagged {\em w}.
\end{enumerate}

\vspace{3mm}

\noindent where {\em a,b,z} and {\em w} are variables over the set of
parts of speech.  To learn a transformation, the learner in essence
applies every possible transformation,\footnote{All possible
instantiations of transformation templates.} counts the number of
tagging errors after that transformation is applied, and chooses that
transformation resulting in the greatest error reduction.\footnote
{The search is data-driven, so only a very small percentage of
possible transformations really need be examined.}  Learning stops
when no transformations can be found whose application reduces errors
beyond some prespecified threshold.  An example of a transformation
that was learned is: change the tagging of a word from {\bf noun} to
{\bf verb} if the previous word is tagged as a {\bf modal}.  Once the
system is trained, a new sentence is tagged by applying the initial
state annotator and then applying each transformation, in turn, to the
sentence.

\section{Lexicalizing the Tagger}

No relationships between words are directly captured in stochastic
taggers.  In the Markov model, state transition probabilities
($P(Tag_{i}|Tag_{i-1} \ldots Tag_{i-n})$) express the likelihood of a
tag immediately following $n$ other tags, and emit probabilities
($P(Word_{j}|Tag_{i})$) express the likelihood of a word given a tag.
Many useful relationships, such as that between a word and the
previous word, or between a tag and the following word, are not
directly captured by Markov-model based taggers.  The same is true of
the earlier transformation-based tagger, where transformation
templates did not make reference to words.

To remedy this problem, the transformation-based tagger was extended
by adding contextual transformations that could make reference to
words as well as part of speech tags.  The transformation templates
that were added are:

\vspace{3mm}

Change tag {\bf a} to tag {\bf b} when:

\begin{enumerate}
\item The preceding (following) word is {\em w}.
\item The word two before (after) is {\em w}.
\item One of the two preceding (following) words is  {\em w}.
\item The current word is {\em w} and the preceding (following) word
is {\em x}.
\item The current word is {\em w} and the preceding (following) word
is tagged {\em z}.
\end{enumerate}

\vspace{3mm}

\noindent where w and x are variables over all words in the training
corpus, and z is a variable over all parts of speech.

Below we list two lexicalized transformations that were
learned:\footnote{All experiments were run on the Penn Treebank tagged
Wall Street Journal corpus, version 0.5 \cite{Marcus93}.}

\vspace{3mm}

\noindent Change the tag:  \\

\noindent(12) From {\bf preposition} to {\bf adverb} if the word two
positions to the right is {\bf as}.\\
\noindent(16) From {\bf non-3rd person singular present verb} to
{\bf base form verb} if one of the previous two words is {\bf
n't}.\footnote{In the Penn Treebank, {\em n't} is treated as a separate
token, so {\em don't} becomes {\em do/VB-NON3rd-SING n't/ADVERB}.} \\

\vspace{3mm}

The Penn Treebank tagging style manual specifies that in the
collocation {\em as \ldots as}, the first {\em as} is tagged as an
adverb and the second is tagged as a preposition.  Since {\em as} is
most frequently tagged as a preposition in the training corpus, the
initial state tagger will mistag the phrase {\em as tall as} as:

\vspace{3mm}

\centerline{as/{\bf preposition} tall/adjective as/preposition}

\vspace{3mm}

The first lexicalized transformation corrects this mistagging.  Note
that a stochastic tagger trained on our training set would not
correctly tag the first occurrence of {\em as}.  Although adverbs are
more likely than prepositions to follow some verb form tags, the fact
that $P(as|preposition)$ is much greater than $P(as|adverb)$, and
$P(adjective|preposition)$ is much greater than $P(adjective|adverb)$
lead to {\em as} being incorrectly tagged as a preposition by a
stochastic tagger.  A trigram tagger will correctly tag this
collocation in some instances, due to the fact that
$P(preposition|\mbox{adverb adjective})$ is greater than
$P(preposition|\mbox{preposition adjective})$, but the outcome will be
highly dependent upon the context in which this collocation appears.

The second transformation arises from the fact that when a verb
appears in a context such as {\em We do n't \_\_\_} or {\em We did n't
usually \_\_\_}, the verb is in base form.  A stochastic trigram
tagger would have to capture this linguistic information indirectly
from frequency counts of all trigrams of the form:\footnote{Where a
star can match any part of speech tag.}

\vspace{3mm}

\begin{center}
\begin{tabbing}
* \hspace{20mm} \= ADVERB \hspace{10mm} \=  PRESENT\_VERB \\

* \> ADVERB \> BASE\_VERB \\

ADVERB \> * \>  PRESENT\_VERB \\

ADVERB \> * \>  BASE\_VERB \\

\end{tabbing}
\end{center}

\vspace{3mm}

\noindent and from the fact that $P(n't|ADVERB)$ is fairly high.

In \cite{Weish93}, results are given when training and testing a
Markov-model based tagger on the Penn Treebank Tagged Wall Street
Journal Corpus.  They cite results making the closed vocabulary
assumption that all possible tags for all words in the test set are
known.  When training contextual probabilities on 1 million words, an
accuracy of 96.7\% was achieved.  Accuracy dropped to 96.3\% when
contextual probabilities were trained on 64,000 words.  We trained the
transformation-based tagger on 600,000 words from the same corpus,
making the same closed vocabulary assumption,\footnote{In both
\cite{Weish93} and here, the test set was incorporated into the
lexicon, but was not used in learning contextual information. Testing
with no unknown words might seem like an unrealistic test.  We have
done so for three reasons (We show results when unknown words are
included later in the paper): (1) to allow for a comparison with
previously quoted results, (2) to isolate known word accuracy from
unknown word accuracy, and (3) in some systems, such as a closed
vocabulary speech recognition system, the assumption that all words
are known is valid.  } and achieved an accuracy of 97.2\% on a
separate 150,000 word test set.  The transformation-based learner
achieved better performance, despite the fact that contextual
information was captured in only 267 simple nonstochastic rules, as
opposed to 10,000 contextual probabilities that were learned by the
stochastic tagger.  To see whether lexicalized transformations were
contributing to the accuracy rate, we ran the exact same test using
the tagger trained using the earlier transformation template set,
which contained no transformations making reference to words.
Accuracy of that tagger was 96.9\%.  Disallowing lexicalized
transformations resulted in an 11\% increase in the error rate.
These results are summarized in table \ref{nounkwds}.

\begin{table}
\begin{center}
\begin{tabular}{|c|c|c|c|} \hline
     &  Training           &  \# of Rules & \\
      &  Corpus &    or Context. &  Acc. \\
Method & Size  (Words)     &  Probs. & (\%) \\ \hline
Stochastic & 64 K     & 6,170         &  96.3 \\ \hline
Stochastic & 1 Million     & 10,000         &  96.7 \\ \hline
Rule-Based  &      &             &   \\
w/o Lex. Rules & 600 K      & 219            &  96.9  \\ \hline
Rule-Based  &      &             &   \\
With Lex. Rules & 600 K      & 267            &  97.2  \\ \hline
\end{tabular}
\end{center}
\caption{Comparison of Tagging Accuracy With No Unknown Words}
\label{nounkwds}
\end{table}

When transformations are allowed to make reference to words and word
pairs, some relevant information is probably missed due to sparse
data.  We are currently exploring the possibility of incorporating
word classes into the rule-based learner in hopes of overcoming this
problem.  The idea is quite simple.  Given a source of word class
information, such as WordNet \cite{Miller90}, the learner is extended
such that a rule is allowed to make reference to parts of speech,
words, and word classes, allowing for rules such as {\em Change the
tag from X to Y if the following word belongs to word class Z}.  This
approach has already been successfully applied to a system for
prepositional phrase disambiguation \cite{Brill93diss}.

\section{Unknown Words}

In addition to not being lexicalized, another problem with the
original transformation-based tagger was its relatively low accuracy
at tagging unknown words.\footnote{This section describes work done in
part while the author was at the University of Pennsylvania.}  In the
initial state annotator for tagging, words are assigned their most
likely tag, estimated from a training corpus.  In the original
formulation of the rule-based tagger, a rather ad-hoc algorithm was
used to guess the most likely tag for words not appearing in the
training corpus.  To try to improve upon unknown word tagging
accuracy, we built a transformation-based learner to learn rules for
more accurately guessing the most likely tag for words not seen in the
training corpus.  If the most likely tag for unknown words can be
assigned with high accuracy, then the contextual rules can be used to
improve accuracy, as described above.

In the transformation-based unknown-word tagger, the initial state
annotator naively labels the most likely tag for unknown words as
proper noun if capitalized and common noun otherwise.\footnote{If we change
the tagger to tag all unknown words as common nouns, then a number of
rules are learned of the form: {\bf change tag to proper noun if the
prefix is "E"}, since the learner is not provided with the concept of
upper case in its set of transformation templates.}

Below we list the set of allowable transformations:

\vspace{3mm}

\noindent Change the tag of an unknown word (from X) to Y if:

\begin{enumerate}
\item Deleting the prefix x, $|x| <= 4$, results in a word (x is any string of
length 1 to 4).
\item The first (1,2,3,4) characters of the word are x.
\item Deleting the suffix x, $|x| <= 4$, results in a word.
\item The last (1,2,3,4) characters of the word are x.
\item Adding the character string x as a suffix results in a word ($|x| <= 4$).
\item Adding the character string x as a prefix results in a word ($|x| <= 4$).
\item Word W ever appears immediately to the left (right) of the word.
\item Character Z appears in the word.
\end{enumerate}

\vspace{3mm}

An unannotated text can be used to check the conditions in all of the
above transformation templates.  Annotated text is necessary in
training to measure the effect of transformations on tagging accuracy.
Below are the first 10 transformation learned for tagging unknown
words in the Wall Street Journal corpus:

\vspace{3mm}

\noindent Change tag:

\begin{enumerate}
\item From {\bf common noun} to {\bf plural common noun} if the
word has suffix {\bf -s}\footnote{Note that this transformation will result
in the mistagging of {\em actress}.  The 17th learned rule fixes this
problem.  This rule states: change a tag from {\bf plural common noun}
to {\bf singular common noun} if the word has suffix {\bf ss}.}
\item From {\bf common noun} to {\bf number} if the word
has character {\bf .}
\item From {\bf common noun} to {\bf adjective} if the word
has character {\bf -}
\item From {\bf common noun} to {\bf past participle verb}
if the word has suffix {\bf -ed}
\item From {\bf common noun} to {\bf gerund or
present participle verb} if the word has suffix {\bf -ing}
\item To {\bf adjective} if adding the suffix {\bf -ly} results
in a word
\item To {\bf adverb} if the word has suffix {\bf -ly}
\item From {\bf common noun} to {\bf number} if the word
{\bf \$} ever appears immediately to the left
\item From {\bf common noun} to {\bf adjective} if the word
has suffix {\bf -al}
\item From {\bf noun} to {\bf base form verb} if the word
{\bf would} ever appears immediately to the left.
\end{enumerate}

\vspace{3mm}

Keep in mind that no specific affixes are prespecified.  A
transformation can make reference to any string of characters up to a
bounded length.  So while the first rule specifies the English suffix
"s", the rule learner also considered such nonsensical rules as: {\em
change a tag to adjective if the word has suffix "xhqr"}.  Also,
absolutely no English-specific information need be prespecified in the
learner.\footnote{This learner has also been applied to tagging Old
English.  See \cite{Brill93diss}.}

We then ran the following experiment using 1.1 million words of the
Penn Treebank Tagged Wall Street Journal Corpus.  The first 950,000
words were used for training and the next 150,000 words were used for
testing.  Annotations of the test corpus were not used in any way to
train the system.  From the 950,000 word training corpus, 350,000
words were used to learn rules for tagging unknown words, and 600,000
words were used to learn contextual rules.  148 rules were learned for
tagging unknown words, and 267 contextual tagging rules were learned.
Unknown word accuracy on the test corpus was 85.0\%, and overall
tagging accuracy on the test corpus was 96.5\%.  To our knowledge,
this is the highest overall tagging accuracy ever quoted on the Penn
Treebank Corpus when making the open vocabulary assumption.

In \cite{Weish93}, a statistical approach to tagging unknown words is
shown.  In this approach, a number of suffixes and important features
are prespecified.  Then, for unknown words:

\vspace{3mm}

$p(W|T) = p(\mbox{unknown word}|T) * p(\mbox{Capitalize-feature}|T)
* p(suffixes,hyphenation|T) $

\vspace{3mm}

Using this equation for unknown word emit probabilities within the
stochastic tagger, an accuracy of 85\% was obtained on the Wall Street
Journal corpus.  This portion of the stochastic model has over 1,000
parameters, with $10^8$ possible unique emit probabilities, as opposed
to only 148 simple rules that are learned and used in the rule-based
approach.  We have obtained comparable performance on unknown words,
while capturing the information in a much more concise and perspicuous
manner, and without prespecifying any language-specific or
corpus-specific information.

\section{K-Best Tags}

There are certain circumstances where one is willing to relax the one
tag per word requirement in order to increase the probability that the
correct tag will be assigned to each word.  In
\cite{DeMarcken90,Weish93}, k-best tags are assigned within a
stochastic tagger by returning all tags within some threshold of
probability of being correct for a particular word.

We can modify the transformation-based tagger to return multiple tags
for a word by making a simple modification to the contextual
transformations described above.  The initial-state annotator is the
tagging output of the transformation-based tagger described above.
The allowable transformation templates are the same as the contextual
transformation templates listed above, but with the action {\em change
tag X to tag Y} modified to {\em add tag X to tag Y} or {\em add tag X
to word W}.  Instead of changing the tagging of a word,
transformations now add alternative taggings to a word.

When allowing more than one tag per word, there is a trade-off between
accuracy and the average number of tags for each word.  Ideally,
we would like to achieve as large an increase in accuracy with as few
extra tags as possible.  Therefore, in training we find
transformations that maximize precisely this function.

In table \ref{kbesttable} we present results from first using the
one-tag-per-word transformation-based tagger described in the previous
section and then applying the k-best tag transformations.  These
transformations were learned from a separate 240,000 word corpus.  As
a baseline, we did k-best tagging of a test corpus as follows.  Each
known word in the test corpus was tagged with all tags seen with that
word in the training corpus and the five most likely unknown word tags
were assigned to all words not seen in the training
corpus.\footnote{Thanks to Fred Jelinek and Fernando Pereira for
suggesting this baseline experiment.}  This resulted in an accuracy of
99.0\%, with an average of 2.28 tags per word.  The rule-based tagger
obtained the same accuracy with 1.43 tags per word, one third the
number of additional tags as the baseline tagger.\footnote{Unfortunately, it is
 difficult to find results to compare
these k-best tag results to.  In \cite{DeMarcken90}, the test set is
included in the training set, and so it is difficult to know how this
system would do on fresh text.  In \cite{Weish93}, a k-best tag
experiment was run on the Wall Street Journal corpus.  They quote the
average number of tags per word for various threshold settings, but do
not provide accuracy results.}

\begin{table}
\begin{center}
\begin{tabular}{|c|c|c|} \hline
\# of Rules &Accuracy & Avg. \# of tags per word \\ \hline
0   & 96.5 & 1.00    \\ \hline
50  & 96.9 & 1.02    \\ \hline
100 & 97.4 & 1.04    \\ \hline
150 & 97.9 & 1.10	     \\ \hline
200 & 98.4 & 1.19	     \\ \hline
250 & 99.1 & 1.50	     \\ \hline
\end{tabular}
\end{center}
\caption{Results from k-best tagging.}
\label{kbesttable}
\end{table}

\section{Conclusions}

In this paper, we have described a number of extensions to previous
work in rule-based part of speech tagging, including the ability to
make use of lexical relationships previously unused in tagging, a new
method for tagging unknown words, and a way to increase accuracy by
returning more than one tag per word in some instances.  We have
demonstrated that the rule-based approach obtains competitive
performance with stochastic taggers on tagging both unknown and known
words. The rule-based tagger captures linguistic information in a
small number of simple non-stochastic rules, as opposed to large
numbers of lexical and contextual probabilities.  Recently, we have
begun to explore the possibility of extending these techniques to
other problems, including learning pronunciation networks for speech
recognition and learning mappings between sentences and semantic
representations.

\end{document}